\documentclass{article}

\usepackage{float}
\usepackage{morefloats}
\usepackage{subfigure}
\usepackage[final]{graphicx}
\usepackage{comment}
\usepackage[utf8]{inputenc} 
\usepackage[T1]{fontenc}    
\usepackage{hyperref}       
\usepackage{url}            
\usepackage{wrapfig}
\usepackage{times}
\usepackage{verbatim}
\usepackage{textcomp} 
\usepackage{multirow}
\usepackage{morefloats}
\usepackage{booktabs}       
\usepackage{amsfonts}
\usepackage{amsthm}
\usepackage{enumitem}       
\usepackage{amsmath}
\usepackage{nicefrac}       
\usepackage{microtype}
\newtheorem{thm}{Theorem}[section]

\theoremstyle{defn}

\theoremstyle{rmk}
\newtheorem{rmk}{Remark}

\newtheorem{pf}{Proof}

\title{On a convergent \textit{off}-policy temporal difference learning algorithm 
in \textit{on}-line learning environment}

\author{Prasenjit Karmakar, Rajkumar Maity, Shalabh Bhatnagar}
\begin{document}

\maketitle

%


\begin{abstract}
 In this paper we provide a rigorous convergence analysis of a ``off''-policy temporal difference learning algorithm 
with linear function approximation and per time-step linear computational complexity in ``online'' learning environment. The algorithm considered here is 
TDC with importance weighting introduced by Maei et al. We support our theoretical results by 
providing suitable empirical results for standard off-policy counterexamples.   
\end{abstract}

\section{Introduction}

We consider the problem of estimating the value function corresponding to a target policy given the 
realization of a finite state Markov decision process under a behaviour policy which is different from the 
target policy. This is well known in literature as the off-policy evaluation problem. 
A solution for this problem might allow one to learn 
about
the optimal policy while behaving according to an exploratory policy. See \cite{suttonb} for 
additional uses.

It is well-known that for this problem the standard temporal difference learning with linear 
function approximation may diverge (\cite{baird}, \cite[Section 3]{emphatic_td}). Further, 
the usual single time-scale stochastic approximation kind of argument may not be useful 
as the associated ordinary differential equation (o.d.e) may not have the TD(0) solution as its globally asymptotically stable 
equilibrium.
In \cite{sutton1,sutton,maeith} the 
gradient temporal difference learning (GTD) algorithms were proposed to solve this problem. The 
per time-step computational complexity for these algorithms scales only linearly in the size $d$ of
the function approximator.
However, the authors 
make the assumption that either 
\begin{enumerate}
 \item one uses ``sub-sampling'' (see \cite[Section 4.1]{maeith},\cite{sutton1} for details) to filter 
the data relevant to target policy given the 
trajectory corresponding to behaviour policy, or
 \item the data itself is available in the off-policy setting i.e. one has direct access 
to quadruples of the form (state, action, reward, next state)
where the first component of the quadruples are sampled independently from 
the stationary distribution of the underlying Markov chain corresponding to the behaviour policy 
and the quadruples are formed according to the target policy.      
\end{enumerate}
Amongst all algorithms with the above assumptions, the TDC (temporal difference learning with gradient correction) algorithm 
was empirically 
found to be most efficient in terms of the rate of convergence. It was shown in \cite{sutton} that such an 
algorithm can be proved to be convergent using the classical convergence proof for two time-scale stochastic approximation with 
martingale difference noise \cite{borkartt}. The reason for using two time-scale framework for the TDC algorithm is to make sure that the 
O.D.E's have globally asymptotically stable equilibrium. However, one can prove the convergence using single time-scale 
convergence analysis as in \cite[Theorem 3]{maeith}; however the extra condition on the step-size ratio $\eta$ 
mentioned there is hard to verify as stationary distribution will be unknown.

Note that such works incorporate the off-policy issue into the data as they don't take the full behvaiour trajectory 
as input to the algorithm. The assumptions used in the aforementioned reference on off-policy algorithms   
are highly restrictive as 
\begin{enumerate}
 \item although in the first case Markov chain sampled at increasing stopping times 
is time-homogeneous 
Markov, its transition probability will be different from the same of the Markov chain corresponding to 
behaviour policy.  
Further, we are interested in an ``online'' learning scheme. Also,
 \item the second situation is not realistic too as the aforementioned stationary distribution will be unknown; 
one has access to only the trajectory corresponding to behaviour policy
from which the goal is to evaluate the target policy.    
\end{enumerate}
Keeping this in mind, another algorithm introduced in \cite{maeith}, namely, TDC with importance weighting solves the above off-policy 
evaluation problem in a more realistic scenario. The idea is to handle the off-policy issue in the algorithm rather than in the data by weighting 
the updates by the likelihood of action taken by the target policy (as opposed to the behavior policy). 
The advantage is that, unlike sub-sampling, here all the data from the given trajectory corresponding 
to the behaviour policy is used which is necessary in an online learning scenario. 
Another advantage of this method is that we can allow both the behaviour and target policies to be
be randomized unlike the sub-sampling scenario
where one can use only deterministic policy as a target policy.
However, to the best of 
our knowledge, both its theoretical and empirical convergence properties have not yet been analyzed. Note that   
one cannot represent the algorithm in the usual two time-scale stochastic approximation  framework to prove its convergence and 
needs to extend such a framework to non-additive Markov noise and additive martingale difference noise. The Markov noise 
appears in the algorithm as the full trajectory of the realization of the  underlying Markov decision process corresponding to 
the behaviour policy is taken as input to the algorithm.

In this work we give a rigorous almost sure convergence  proof of TDC algorithm with importance weighting by formulating 
it into the two time-scale stochastic approximation framework with non-additive Markov noise and additive martingale 
difference noise. 
To the best of our knowledge this 
is the first time an almost sure convergence proof of off-policy temporal difference learning algorithm with linear function
approximation is presented for step-sizes satisfying Robbins-Monro conditions. We also support these theoretical results 
by providing empirical results. Our results show that due to 
the above-mentioned importance weighting factor, 
online TDC with importance weighting performs 
much better than the sub-sampling version of TDC for standard off-policy counterexamples  
when the behaviour policy is much different from the 
target policy. 

Recently, emphatic temporal difference learning has been introduced in \cite{emphatic_td} to solve the off-policy evaluation 
problem. 
However, such algorithms are proven to be almost surely convergent for special step-size sequences and 
weakly convergent for a large range of step-sizes \cite{yu_new}. 

Another related work is the much complex off-policy learning algorithms that obtain
the benefits of weighted importance sampling (to reduce variance)  with $O(d)$ computational
complexity \cite{weight}. However, nothing is known about the convergence of such algorithms.
In this context, we empirically show that in the case of TDC with importance weighting 
the variance of the difference between true value 
function and the estimated one for standard off-policy counterexamples such as \cite{baird} becomes 
very small eventually.  

The organization of the paper is as follows: Section \ref{back} describes the TDC algorithm with importance weighting. 
Section \ref{main_res} gives the rigorous convergence proof of the algorithm. Section \ref{empiric} 
shows empirical results supporting our theoretical results. 
Finally we conclude by providing some interesting future directions. 
  
\section{Background and description of TDC with importance weighting}
\label{back}
We need to estimate the value function for a target policy $\pi$ 
given the continuing evolution of the underlying MDP (with finite state and action spaces $S$ and $A$ respectively,   
specified by expected reward $r(\cdot,\cdot,\cdot)$ 
and transition probability kernel $p(\cdot|\cdot,\cdot)$) for a 
behaviour policy $\pi_b$ with $\pi \neq \pi_b$. 
Suppose, the above-mentioned on-policy trajectory is $(X_{n},A_n, R_{n},X_{n+1}), n\geq 0$ where  
$\{X_n\}$ is a time-homogeneous irreducible Markov chain with unique stationary distribution $\nu$ 
and generated from the behavior policy $\pi_b$. Here the quadruplet $(s,a,r,s')$ 
represents (current state, action, reward, next state). 
Also, assume that $\pi_b(a|s) > 0 ~\forall s \in S, a \in A$. We need 
to find the solution $\theta^*$ for the following: 
\begin{equation}
\label{fixpoint}
\begin{split}
0&=\sum_{s,a,s'}\nu(s)\pi(a|s)p(s'|s,a)\delta(\theta;s,a,s')\phi(s) = E[\rho_{X,A}\delta_{X,R,Y}(\theta)\phi(X)]\\ 
 &= b - A\theta,
\end{split}
\end{equation}
where 
\begin{itemize}
\item [(i)]$\theta \in \mathbb{R}^d$ is the parameter for value function,
\item [(ii)]$\phi: S\to \mathbb{R}^d$ is a vector of state features,
\item [(iii)] $X \sim \nu$,
\item [(iv)] $0<\gamma < 1$ is the discount factor,
\item [(v)] $E[R|X=s,Y=s'] = \sum_{a\in A}\pi_b(a|s)r(s,a,s')$,
\item [(vi)] $P(Y=s'|X=s) = \sum_{a\in A}\pi_b(a|s) p(s'|s,a)$,
\item [(vii)] $\delta(\theta; s,a,s')= r(s,a,s') + \gamma \theta^T\phi(s') - \theta^T\phi(s)$
is the temporal difference term with expected single-stage reward,
\item [(viii)] $\rho_{X,A} = \frac{\pi(A | X)}{\pi_b(A|X)}$,
\item [(ix)] $\delta_{X,R,Y}=R + \gamma \theta^T\phi(Y) - \theta^T\phi(X)$,
\item [(x)] $A=E[\rho_{X,A}\phi(X)(\phi(X) -\gamma\phi(Y))^T]$, $b=E[\rho_{X,A}R\phi(X)]$.
\end{itemize}
The desired approximate value function under the target policy $\pi$ is $V_{\pi}^*={\theta^*}^T\phi$. 
Let $V_\theta = {\theta}^T\phi$.
It is well-known (\cite{maeith}) that $\theta^*$ (solution to (\ref{fixpoint})) satisfies the projected fixed point equation namely
\begin{equation}
V_{\theta}= \Pi_{\mathcal{G},\nu}T^{\pi}V_{\theta},\nonumber 
\end{equation}
where 
\begin{equation}
\Pi_{\mathcal{G}, \nu}\hat{V} =
\arg\min_{f \in \mathcal{G}} (\|\hat{V} - f\|_{\nu}),\nonumber 
\end{equation}
with $\mathcal{G} = \{V_{\theta} | \theta \in \mathbb{R}^d\}$
and the Bellman operator
\begin{equation}
T^{\pi}V_\theta(s) = \sum_{s' \in S} \sum_{a\in A}\pi(a|s)p(s'|s, a)\left[\gamma V_\theta(s') + r(s, a, s')\right]. \nonumber 
\end{equation}
Here $\|\cdot\|_{\nu}$ is the weighted Euclidean norm defined by $\|f\|^2_{\nu}=\sum_{s\in S}f(s)^2 \nu(s)$,
Therefore to find $\theta^*$, the idea is to minimize the mean square projected 
Bellman error (MSPBE) $J(\theta)= \|V_{\theta} - \Pi_{\mathcal{G},\nu}T^{\pi}V_{\theta}\|^2_{\nu}$ using stochastic gradient descent.
It can be shown that the expression of gradient contains product of multiple expectations. Such framework can be modelled by 
two time-scale stochastic approximation where one iterate stores the quasi-stationary estimates of some of the expectations and the 
other iterate is used for sampling. 

%

We consider the TDC (Temporal Difference with Correction) algorithm with importance-weighting 
from Sections 4.2 and 5.2 of \cite{maeith}. 
The gradient in this case can be shown to satisfy 
\begin{align}
-\frac{1}{2}\nabla J(\theta)&=E[\rho_{X,A}\delta_{X,R,Y}(\theta)\phi(X)] - \gamma E[\rho_{X,A}\phi(Y)\phi(X)^T]w(\theta),\nonumber\\
w(\theta) &= E[\phi(X)\phi(X)^T]^{-1}E[\rho_{X,A}\delta_{X,R,Y}(\theta)\phi(X)].\nonumber 
\end{align}Define $\phi_n = \phi(X_n)$, $\phi'_n = \phi(X_{n+1})$, $\delta_n(\theta) = \delta_{X_n, R_n, X_{n+1}}(\theta)$ and 
$\rho_n=\rho_{X_n,A_n}$.
Therefore the associated iterations in this algorithm are: 
\begin{align}
\theta_{n+1} &= \theta_n + a(n) \rho_n\left[\delta_{n}(\theta_n)\phi_n - \gamma \phi'_{n}\phi_n^Tw_n\right],\label{tdc_slow} \\
w_{n+1} &= w_n + b(n) \left[(\rho_n\delta_{n}(\theta_n) - \phi_n^Tw_n)\phi_n\right], \label{tdc_fast}
\end{align}
\\ \indent 
with $\{a(n)\}, \{b(n)\}$ satisfying conditions which will be specified later. 
Note that the second term inside bracket in (\ref{tdc_slow}) is essentially an adjustment or correction
of the TD update so that it follows the gradient of the
MSPBE objective function thus helping in the desired convergence.

Note that the sub-sampling version of TDC algorithm (therefore the offline version of TDC
algorithm) can be written in the following way:
\begin{align}
\theta_{n+1} &= \theta_n + a(n) I_{\{A_n = \pi(X_n)\}}\left[\delta_{n}(\theta_n)\phi_n - \gamma \phi'_{n}\phi_n^Tw_n\right],\nonumber \\
w_{n+1} &= w_n + b(n) I_{\{A_n = \pi(X_n)\}}\left[(\delta_{n}(\theta_n) - \phi_n^Tw_n)\phi_n\right], \nonumber
\end{align}   
where $I_{\{A_n = \pi(X_n)\}} =1$ if $A_n = \pi(X_n)$ and $0$ otherwise.
In the rest of the paper both the above algorithms will be denoted by ONTDC and OFFTDC respectively except the 
figures in Section \ref{empiric} where we mention the full name.

\section{Almost sure convergence proof of ONTDC}
\label{main_res}
As mentioned earlier, to analyze the convergence of the iterations (\ref{tdc_slow}) and (\ref{tdc_fast}) one has to first extend 
the classic two time-scale stochastic approximation framework of Borkar \cite{borkartt} to a setting with  
Markov noise. The full extension is 
shown in the Appendix. We only state here a special case of this theory which will be sufficient for us.
Hence we start with this extension and 
then later show how the TDC iterations can be cast into this framework and proven to be convergent. 
\subsection{Two timescale stochastic approximation with Markov noise}
\label{sec_def}
Our goal is to perform an asymptotic analysis of the following coupled recursions:  
\begin{eqnarray}
\theta_{n+1} &= \theta_n + a(n)\left[h(\theta_n, w_n, Z^{(1)}_n) + M^{(1)}_{n+1}\right],\label{eqn1}\\
w_{n+1} &= w_n + b(n)\left[g(\theta_n, w_n, Z^{(2)}_n) + M^{(2)}_{n+1}\right],\label{eqn2}
\end{eqnarray}
where $\theta_n \in \mathbb{R}^d, w_n \in \mathbb{R}^k, n\geq 0$ and $\{Z^{(i)}_n\}, \{M^{(i)}_{n}\}, i=1, 2$ 
are random processes that we describe below. 
\\ \indent 
We make the following assumptions:  
\begin{enumerate}[label=\textbf{(A\arabic*)}]
 \item $\{Z^{(i)}_n\}$ takes values in a compact metric space $S^{(i)}, i=1,2$. Additionally, 
the processes $\{Z^{(i)}_n\}, i = 1, 2$ are  
Markov processes  
with their individual dynamics specified by
\begin{equation}
P(Z^{(i)}_{n+1} \in B^{(i)} |Z^{(i)}_m, m\leq n) = \int_{B^{(i)}} p^{(i)}(dy|Z^{(i)}_n), n\geq 0, \nonumber 
\end{equation} 
for $B^{(i)}$ Borel in $S^{(i)}, i = 1, 2,$ respectively.  

\item $h :  \mathbb{R}^{d+k} \times S^{(1)} \to \mathbb{R}^d$ is  
jointly continuous as well as Lipschitz in its first two arguments uniformly w.r.t the third. The latter condition means that
\begin{equation}
\forall z^{(1)} \in S^{(1)}, \|h(\theta, w, z^{(1)}) - h(\theta', w', z^{(1)})\| \leq L^{(1)}(\|\theta-\theta'\| + \|w - w'\|).\nonumber
\end{equation}
Same thing is also true for $g$ where the Lipschitz constant is $L^{(2)}$.
Note that the Lipschitz constant $L^{(i)}$ does not depend on $z^{(i)}$ for $i=1,2$.

\item $\{M^{(i)}_n\}, i=1, 2$ are martingale difference sequences 
w.r.t the increasing $\sigma$-fields
\begin{equation}
\mathcal{F}_n = \sigma(\theta_m, w_m, M^{(i)}_{m}, Z^{(i)}_m, m \leq n, i = 1, 2), n \geq 0,\nonumber 
\end{equation}
satisfying 
\begin{equation}
E[\|M^{(i)}_{n+1}\|^2|\mathcal{F}_n] \leq K(1 + \|\theta_n\|^2 + \|w_n\|^2), i = 1, 2,\nonumber 
\end{equation}
for $n \geq 0$ and a given constant $K>0$.
\item The stepsizes $\{a(n)\}, \{b(n)\}$ are positive scalars satisfying
\begin{equation}
\sum_n a(n) = \sum_n b(n) = \infty, \sum_{n}(a(n)^2 + b(n)^2) < \infty, \frac{a(n)}{b(n)} \to 0.\nonumber 
\end{equation}
Moreover, $a(n), b(n), 
n \geq 0$ are non-increasing. 
\item The map $S^{(i)} \ni z^{(i)}  
\to p^{(i)}(dy|z^{(i)}) \in \mathcal{P}(S^{(i)})$ is continuous. 
\item The function $\hat{g}(\theta, w) = \int g(\theta, w, z)\Gamma^{(2)}(dz)$ is Lipschitz 
continuous where $\Gamma^{(2)}$ is the unique stationary distribution of $Z^{(2)}_n$. 
Further, for all $\theta \in \mathbb{R}^d$, the o.d.e
\begin{equation}
\label{fast_ode}
\dot{w}(t) = \hat{g}(\theta, w(t)) 
\end{equation}
has globally asymptotically stable equilibrium $\lambda(\theta)$ 
where $\lambda :  \mathbb{R}^d \to \mathbb{R}^k$ is a Lipschitz map with constant $K$.
Moreover, the function $V': \mathbb{R}^{d+k} \to 
[0,\infty)$ defined by $V'(\theta,w) = V_\theta(w)$ is continuously differentiable where $V_\theta(.)$ is the Lyapunov function 
for $\lambda(\theta)$. 
This extra condition is needed
so that the set graph($\lambda$):=$\{(\theta,\lambda(\theta)): \theta \in \mathbb{R}^d\}$ becomes a 
globally asymptotically stable set of the coupled o.d.e 
\begin{equation}
\dot{w}(t) = \hat{g}(\theta(t),w(t)), \dot{\theta}(t) = 0.\nonumber  
\end{equation} 
\item Let $\hat{h}(\theta)=\int h(\theta,\lambda(\theta),z) \Gamma^{(1)}(dz)$ 
where $\Gamma^{(1)}$
 is the unique stationary distribution of the Markov process $Z^{(1)}$. Then the o.d.e
\begin{equation}
\label{slow_ode}
\dot{\theta}(t) = \hat{h}(\theta(t))), 
\end{equation}
has a globally asymptotically stable equilibrium $\theta^*$.  
\item Stability of the iterates: $\sup_n(\|\theta_n\| + \|w_n\|) < \infty$ a.s.
\end{enumerate}

The following theorem is our main result:
\begin{thm}[Slower timescale result]Under assumptions \textbf{(A1)-(A8)}, 
\label{main_thm}
\begin{equation}
(\theta_n, w_n) \to (\theta^*, \lambda(\theta^*)) \mbox{a.s. as $n \to \infty$.}\nonumber 
\end{equation}
\end{thm}
We call (\ref{fast_ode}) and (\ref{slow_ode}) as the faster and slower o.d.e 
to correspond with faster and slower recursions, respectively.   

\subsection{Convergence Proof of ONTDC}

\begin{thm}
\label{th2}
Consider the iterations (\ref{tdc_slow}) and (\ref{tdc_fast}) of the TDC. Assume the following:
\begin{enumerate}[label=(\roman*)]
 \item $\{a(n)\}, \{b(n)\}$ satisfy \textbf{(A4)}.
 \item $\{(X_n,R_n,X_{n+1}), n\geq0\}$ is such that $\{X_n\}$ is a time-homogeneous finite state irreducible Markov chain 
   generated from the behavior policy $\pi_b$ with unique stationary distribution $\nu$. 
$E[R_{n}|X_{n}=s,X_{n+1}=s'] = \sum_{a\in A} \pi_b(a|s)r(s,a,s')$ and $P(X_{n+1} =s'|X_{n}=s) = \sum_{a\in A}\pi_b(a|s)p(s'|s,a)$ 
where $\pi_b$ is the behaviour  policy,  
$\pi \neq \pi_b$.  
Also, $E[R_n^2 | X_n, X_{n+1}] < \infty$ for all $n$ almost surely, and
\item $C=E[\phi(X)\phi(X)^T]$ and  $A=E[\rho_{X,A}\phi(X)(\phi(X) -\gamma\phi(Y))^T]$ are non-singular where $X \sim \nu$. 
\item $\pi_b(a|s) > 0$ for all $s \in S, a \in A$.
\item $\sup_n(\|\theta_n\| + \|w_n\|) < \infty$ w.p. 1. 
\end{enumerate}
Then the parameter vector $\theta_n$ converges
with probability one as $n \to \infty$ to the TD(0) solution (\ref{fixpoint}). 
\end{thm}
\begin{pf}
The iterations (\ref{tdc_slow}) and (\ref{tdc_fast}) can be cast into the framework of Section \ref{sec_def} 
with \begin{enumerate}[label=(\roman*)]
\item $Z^{(i)}_n = X_{n-1}$,
\item $h(\theta,w,z) = E[(\rho_n(\delta_{n}(\theta_n)\phi_n-\gamma \phi'_{n}\phi_n^Tw_n))|X_{n-1}=z,\theta_n=\theta,w_n=w]$,
\item $g(\theta,w,z)=E[((\rho_n\delta_{n}(\theta_n) - \phi_n^Tw_n)\phi_n)|X_{n-1}=z,\theta_n=\theta,w_n=w]$,
\item $M^{(1)}_{n+1}=\rho_n(\delta_{n}(\theta_n)\phi_n - \gamma \phi'_{n}\phi_n^Tw_n)-E[\rho_n(\delta_{n}(\theta_n)\phi_n - \gamma \phi'_{n}\phi_n^Tw_n)|X_{n-1}, \theta_n, w_n]$,
\item $M^{(2)}_{n+1}=(\rho_n\delta_{n}(\theta_n) - \phi_n^Tw_n)\phi_n - E[(\rho_n\delta_{n}(\theta_n) - {\phi_n}^T w_n)\phi_n|X_{n-1}, \theta_n, w_n]$,
\item $\mathcal{F}_n = \sigma(\theta_m, w_m, R_{m-1}, X_{m-1},A_{m-1}, m \leq n, i = 1, 2), n \geq 0$. 
\end{enumerate}
Note that in (ii) and (iii) we can define $h$ and $g$ independent of $n$ due to time-homogeneity of $\{X_n\}$.  
\\ \indent
Now, we 
verify the assumptions \textbf{(A1)-(A8)} (mentioned in Section \ref{sec_def}) for our application:
\begin{enumerate}[label=(\roman*)]
\item  \textbf{(A1)}: $Z^{(i)}_n, \forall n, i=1,2$ takes values in compact metric space as $\{X_n\}$ is a finite state Markov chain.  
\item  \textbf{(A5)}: Continuity of transition kernel follows trivially from the fact that we have a finite state MDP.
 \item \textbf{(A2)}\begin{align} &\|h(\theta, w,z) - h(\theta',w',z)\|\nonumber\\
        &=\|E[\rho_n(\theta-\theta')^T(\gamma \phi(X_{n+1}) - \phi(X_n))\phi(X_n) \nonumber \\
         &- \gamma \rho_n \phi(X_{n+1})\phi(X_n)^T(w-w')|X_{n-1}=z]\|\nonumber\\
        &\leq L(2\|\theta-\theta'\|M^2 + \|w-w'\|M^2)\nonumber,
       \end{align}
where $M=\max_{s\in S}\|\phi(s)\|$ with $S$ being the state space of the MDP and $L=\max_{(s,a)\in (S\times A)}\frac{\pi(a|s)}{\pi_b(a|s)}$. 
Hence $h$ is Lipschitz continuous in the 
first two arguments uniformly w.r.t the third. In the last inequality above, we use the 
Cauchy-Schwarz inequality. As with the case of $h$, $g$ can be shown to be Lipschitz continuous in the 
first two arguments uniformly w.r.t the third. Joint continuity of $h$ and $g$ follows from the above
as well as the finiteness of $S$.
\item \textbf{(A3)}: Clearly, $\{M_{n+1}^{(i)}\}, i=1,2$ are martingale difference sequences w.r.t. increasing $\sigma$-fields $\mathcal{F}_n$.
Note that $E[\|M_{n+1}^{(i)}\|^2 | \mathcal{F}_n] \leq K(1 + \|\theta_n\|^2 + \|w_n\|^2)$ a.s., $n\geq 0$ since
$E[R_n^2 | X_n, X_{n+1}] < \infty$ for all $n$ almost surely and $S$ is finite.
\item \textbf{(A4)}: This follows from the conditions (i) in the statement of Theorem \ref{th2}.  
\end{enumerate}

Now, one can see that 
the faster o.d.e. becomes 
\begin{equation}
\dot{w}(t)=E[\rho_{X,A}\delta_{X,R,Y}(\theta)\phi(X)] - E[\phi(X)\phi(X)^T]w(t).\nonumber 
\end{equation}
Clearly, $C^{-1}E[\rho_{X,A}\delta_{X,R,Y}(\theta)\phi(X)]$
is the globally asymptotically stable equilibrium of the o.d.e. The corresponding Lyapunov function 
$V(\theta,w) = \frac{1}{2} \|Cw - E[\rho_{X,A}\delta_{X,R,Y}(\theta)\phi(X)]\|^2$ is 
continuously differentiable. Additionally, $\lambda(\theta)=C^{-1}E[\rho_{X,A}\delta_{X,R,Y}(\theta)\phi(X)]$ 
and it is   
Lipschitz continuous in $\theta$, verifying \textbf{(A6)}. , Further, $A^{-1}E[\rho_{X,A}R\phi(X)]$ is the globally
asymptotically stable equilibrium of the slower o.d.e., verifying \textbf{(A7)}.  
Also, \textbf{(A8)} is (v) in the statement of Theorem \ref{th2}. 
Therefore 
the assumptions $(\mathbf{A1}) - (\mathbf{A8})$ are verified. The proof then follows
from Theorem \ref{main_thm}. 
\end{pf}

\begin{rmk}
Because of the fact that the gradient is a product of two expectations the scheme 
is a  ``pseudo''-gradient descent which helps to find 
the global minimum here.
\end{rmk}
\begin{rmk}
Here we assume the stability of the iterates (\ref{tdc_slow}) and (\ref{tdc_fast}). 
Certain sufficient conditions have been sketched for showing  
stability of single timescale stochastic recursions with controlled Markov noise 
\cite[p.~75, Theorem 9]{borkar1}. This subsequently needs to be 
extended to the case of two time-scale recursions. In this context we mention that 
the way single timescale Borkar-Meyn theorem was used in \cite{sutton} to prove stability 
of two time-scale recursions is not a proper way to prove the same.
\end{rmk}
\begin{rmk}
Convergence analysis for ONTDC along with 
eligibility traces cf. \cite[p.~74]{maeith} where it is called GTD($\lambda$) can be done similarly using our results. 
The main advantage is that it works for $\lambda < \frac{1}{L\gamma}$ ($\lambda\in [0,1]$ being the eligibility function) 
whereas the analysis in \cite{yu} is shown  
only for $\lambda$ very close to 1.   
\end{rmk}

\section{Empirical results}
\label{empiric}
For the assessment of the algorithm experimentally we have compared the result 
on a variation of the classic Baird's off-policy counter-example \cite[Fig. 2.4]{maeith} 
and $\theta\rightarrow 2\theta$ problem \cite[Section 3]{emphatic_td}. In 
both cases, we compare the TD(0), OFFTDC and ONTDC. Unlike \cite{sutton} 
where updating was done synchronously
in dynamic-programming-like sweeps through the
state space, we consider the usual stochastic approximation scenario where only simulated sample trajectories 
are taken as input to the algorithms i.e. the algorithms do not use any knowledge of the probabilities 
for the underlying Markov decision process. For Baird's problem our performance metric is 
Root Mean Squared Error (RMSE) defined to be  
the square root of the average of the square of the deviation between 
true value function and the estimated value function. For $\theta \to 2\theta$ problem 
the $y$-axis is $\theta$ itself.
The average is taken over 1000 simulation run and the metric 
is plotted  against the number of times 
$\theta_n$ is updated. While 
the analysis has been shown for the diminishing step-size case, we implement here the 
algorithm with constant step-sizes as in \cite{maeith,sutton}. 

\begin{figure}
  \centering
  \begin{minipage}[b]{0.49\textwidth}
    \includegraphics[width=\textwidth]{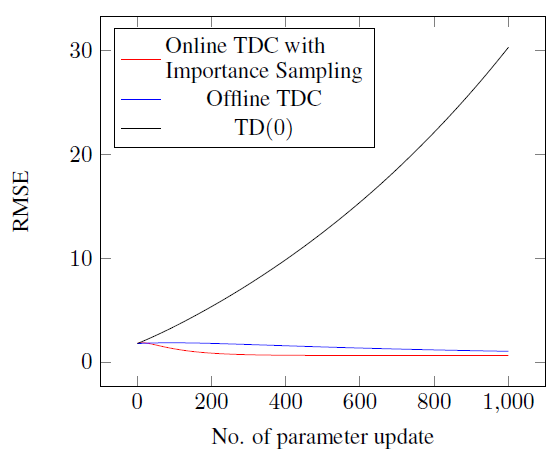}
    \caption{Comparison between TD(0), OFFTDC and ONTDC for Baird's counterexample}
\label{baird_off}  
\end{minipage}
  \hfill
\begin{minipage}[b]{0.49\textwidth}
    \includegraphics[width=\textwidth]{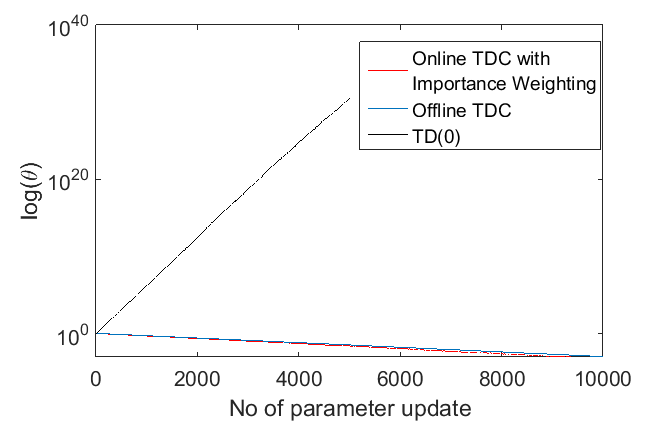}
    \caption{Comparison between TD(0), OFFTDC and ONTDC for $\theta \to 2 \theta$}
\label{theta_2}  
\end{minipage}
\end{figure}

The  $\theta\rightarrow 2\theta$ problem consists of only 2 states where $\theta$ 
and $2\theta$ are the estimated value of the states. According to its behavior policy 
with probability $p=\frac{1}{2}$ it stays on the same state and chooses the other 
state. The target policy is to choose the action that accesses the second 
state with probability 1 (See Fig. 1 in \cite[Section 3]{emphatic_td} for details). 
The constant step-sizes  are chosen as $a(n)=.075;b(n)=.05$ for the two time-scale algorithms and 
$\alpha=.075$ for single timescale algorithms. The simulations are run for 1000 
different sample paths. Rewards in all transitions 
are zero. The initial values are $\theta=1 $ and $w=0$. The results are summarized in Figure \ref{theta_2}.

\begin{figure}
\centering
\subfigure[$p=.01$]{\label{}\includegraphics[width=.49\linewidth]{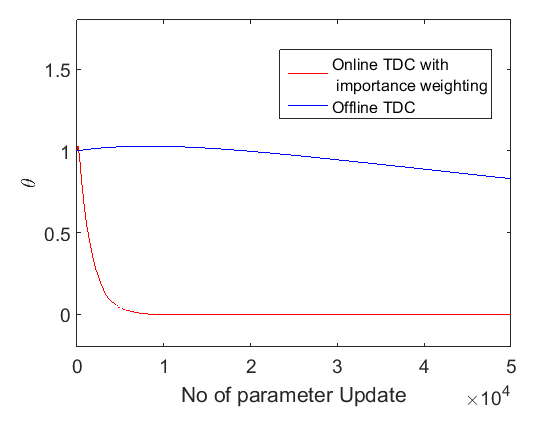}}\hfill
\subfigure[$p=.001$]{\label{}\includegraphics[width=.49\linewidth]{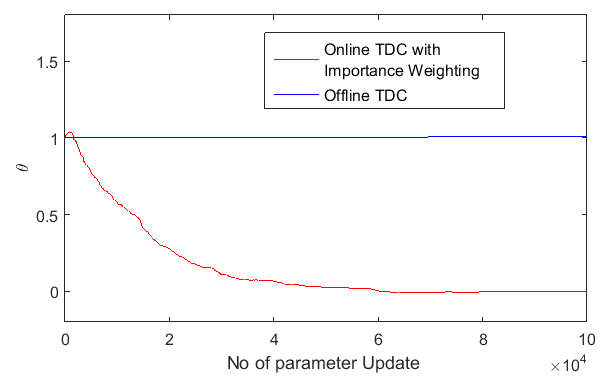}}
\caption{Comparison of OFFTDC and ONTDC for $\theta \to 2\theta$ problem for different values of $p$}
\label{offvson}
\end{figure}
\begin{figure}
\centering
\subfigure[$q=.01$]{\label{}\includegraphics[width=.49\linewidth]{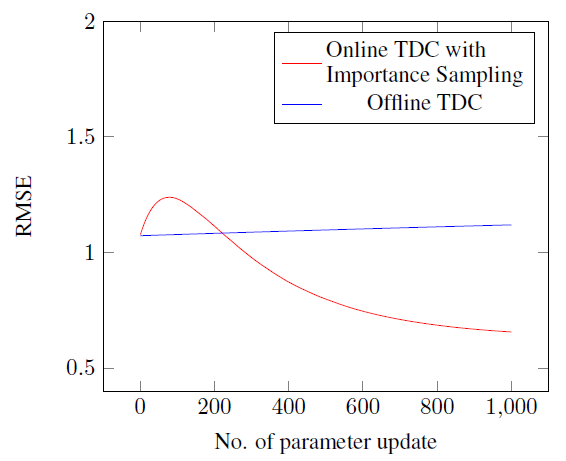}}\hfill
\subfigure[$q=.001$]{\label{}\includegraphics[width=.49\linewidth]{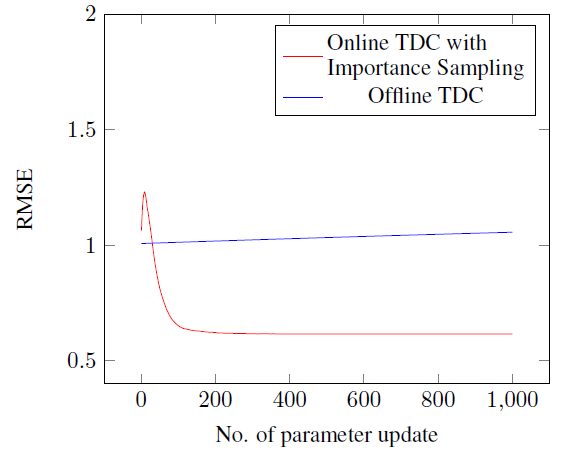}}
\caption{Comparison of OFFTDC and ONTDC for Baird's counterexample for different values of $q$}
\label{offvsonb}
\end{figure}
Next we consider the '7-star' version of Baird's counter example from \cite[p~.17]{maeith} 
All the rewards in transitions are zero and true value function for
each state is zero. The value functions are approximated as 
$V(s)=2\theta(s)+\theta_0 $ $\forall s \in \{1,2 \ldots 6\}$ and 
$V(7)=\theta(7)+ 2\theta_0$. The behaviour policy is to 
choose the state $7$ with probability $q=\frac{1}{7}$ and choose 
uniformly states $1-6$ with probability $(1-q)=\frac{6}{7}$. 
The target policy is to choose the state $7 $ with probability 1. 
The step size chosen for this setting is $a=.005,b=.05$. The 
initial parameters are $\theta=(1,1,1,1,1,1,10,1)$ and $w=\mathbf{0}$. The results in this 
case are summarized in Figure \ref{baird_off}. 
\begin{figure}
 \centering
\includegraphics[width=3 in]{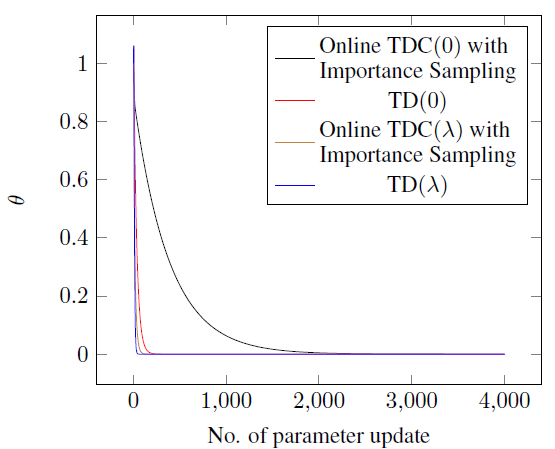}
\caption{on-policy learning on $\theta \to 2\theta$ problem}
\label{theta_on}   
\end{figure}

In both cases (Fig. \ref{theta_2} and \ref{baird_off}) ONTDC performs better than the OFFTDC. 
The difference becomes more apparent when
behaviour policy differs significantly from the target policy (Fig \ref{offvson} and \ref{offvsonb}). The intuition is that in case of OFFTDC
the TD update is weighted by only step-size whereas in case of ONTDC it is 
additionally weighted by $\rho_n$. Therefore by changing the behaviour policy one can improve the rate of convergence 
of the algorithm. In the case  of on-policy learning for 
the $\theta \to 2\theta$ problem, Figure \ref{theta_on} shows that 
with eligibility traces the performance of ONTDC is much closer to $TD(\lambda)$ compared to 
the case with $\lambda=0$.

\begin{figure}
\centering
\subfigure[Baird's counterexample]{\label{fig5:a}\includegraphics[width=.49\linewidth]{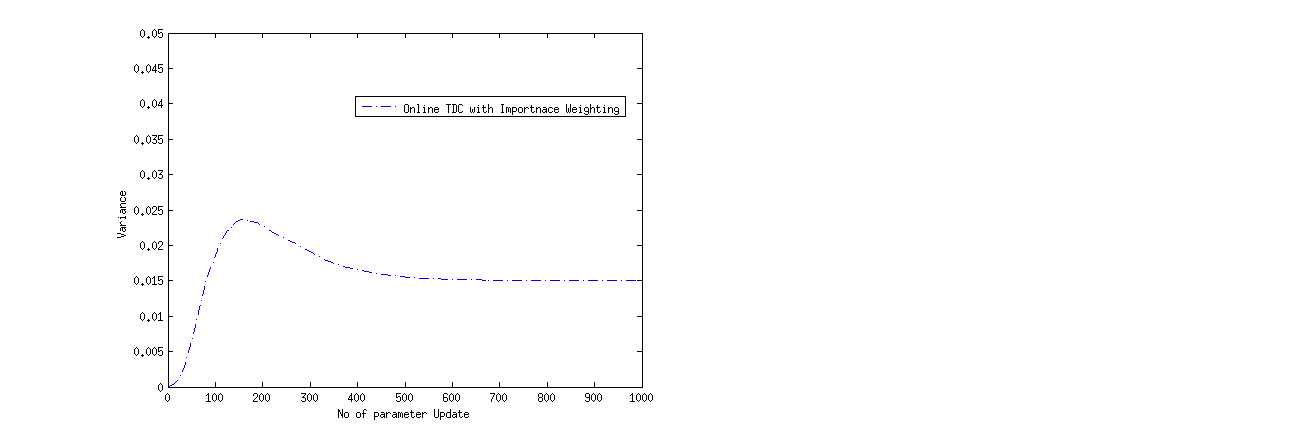}}\hfill
\subfigure[ $\theta \to 2\theta$ problem]{\label{fig5:a}\includegraphics[width=.49\linewidth]{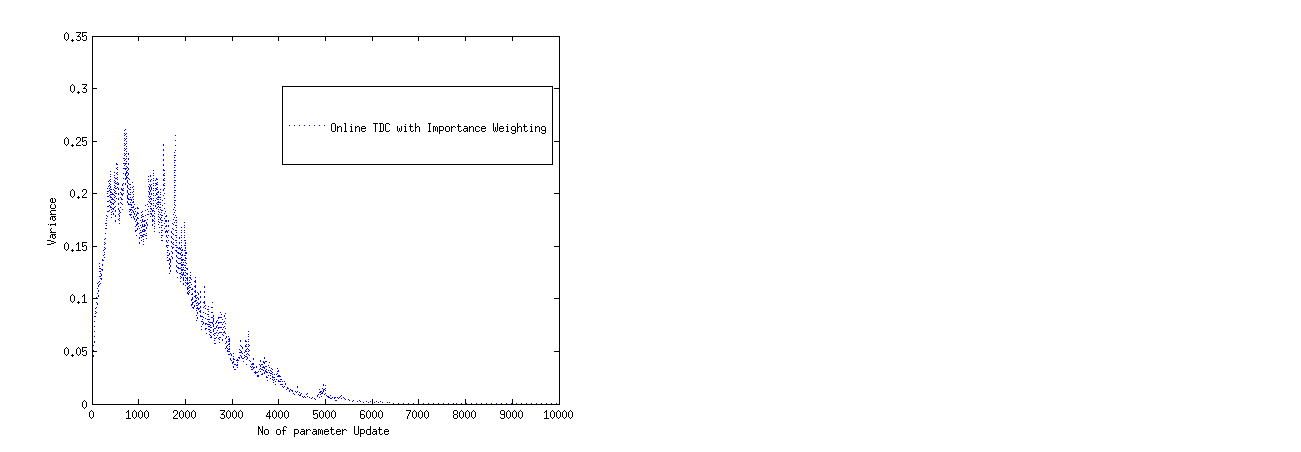}}
\caption{Variances of the performance metric for ONTDC}
\label{var}
\end{figure}

Although ONTDC uses importance weighting in its update, this is not importance sampling used in Monte-Carlo 
algorithms which is the source of high variance. Further, ONTDC 
does not have any follow-on trace like emphatic TD which has a high variance. 
We show in Fig. \ref{var} that the variances of the performance metric for the 
ONTDC is negligible 
eventually for the two standard counterexamples. 

\begin{figure}
\centering
\subfigure[Baird counterexample]{\label{ebaird}\includegraphics[width=.49\linewidth]{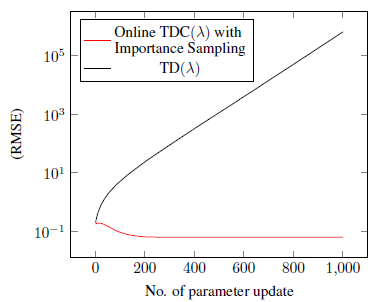}}\hfill
\subfigure[$\theta \to 2\theta$ problem]{\label{etheta}\includegraphics[width=.49\linewidth]{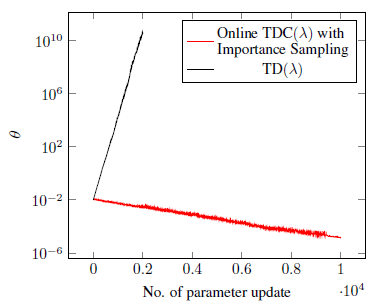}}
\caption{Comparison of TD($\lambda$) and TDC($\lambda$)}
\label{eligib}
\end{figure}

\begin{figure}
\centering
\subfigure[Baird counterexample]{\label{dimstepb}\includegraphics[width=.49\linewidth]{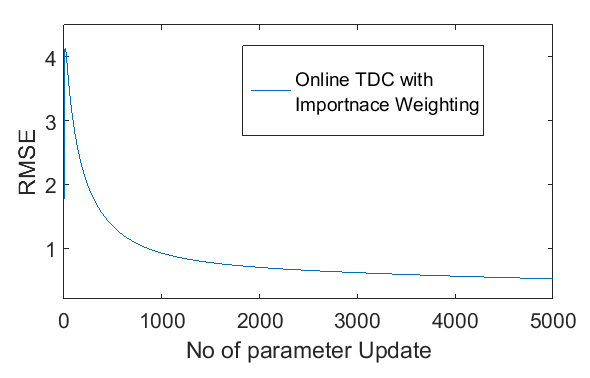}}\hfill
\subfigure[$\theta \to 2\theta$ problem]{\label{dimstept}\includegraphics[width=.49\linewidth]{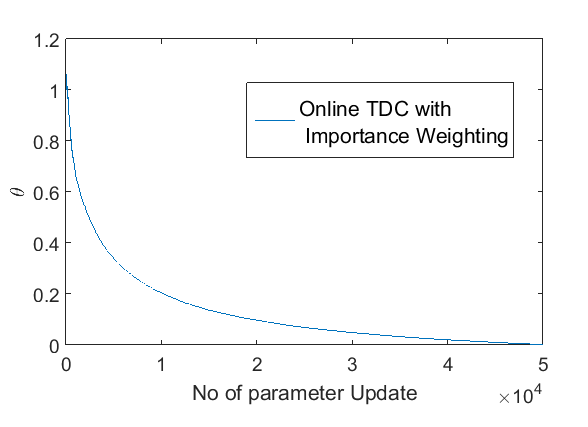}}
\caption{ONTDC with step size for (a) Baird's counterexample $a(n)=\frac{.5}{n}$,
$b(n)=\frac{.125}{n^{.95}}$ and (b) $\theta \to 2\theta$ problem $a(n)=\frac{7}{n+100}$,
$b(n)=\frac{.5}{n^{.95}}$.}
\label{dimstep}
\end{figure}

For both the aforementioned examples the results for the extension to eligibility traces (the algorithm is called GTD($\lambda$)
or TDC($\lambda$)) 
can be seen in Fig. \ref{eligib} with $\lambda =0.1$.
Fig. \ref{dimstep} shows the results of experiments  
where the step-size sequences obey the requirements in \textbf{(A5)}. We observe good convergence 
behaviour in this case that is also better when compared with the case of constant step-sizes as considered in the main paper. 

\section{Conclusion}
We presented almost sure 
convergence proof for an off-policy temporal difference learning algorithm that is also 
extendible to eligibility traces (for a sufficiently large 
range of $\lambda$) with linear function approximation under the assumption 
that the ``on-policy'' trajectory for a behaviour policy is only available. 
This has previously not been done to our knowledge. 

A future direction would be to similarly extend algorithms for off-policy control (\cite{greedygq}) to the more 
realistic settings as we consider in this paper.

\newpage
\small
\bibliographystyle{plain} 
\bibliography{mybib}

\end{document}